\documentclass[12pt]{article}

\usepackage{scicite}
\usepackage{amsmath}

\usepackage{makecell}
\usepackage{textcomp}
\usepackage{graphics}
\usepackage{epsfig}
\usepackage{mathptmx}
\usepackage{times}
\usepackage{amsmath}
\usepackage{amssymb}
\usepackage{multirow}
\usepackage{hhline}
\usepackage{booktabs}
\usepackage{float}
\usepackage{array}
\usepackage{makecell}
\usepackage{lipsum}
\usepackage{caption}
\usepackage{subcaption}
\usepackage{hyperref}
\usepackage{soul}
\usepackage[export]{adjustbox}

\DeclareMathOperator*{\argmax}{arg\,max}
\DeclareMathOperator*{\argmin}{arg\,min}
\usepackage{times}
\usepackage[symbol]{footmisc}

\usepackage{lipsum}

\DefineFNsymbols*{lamportnostar}[math]{\dagger\ddagger\S\P\|{\dagger\dagger}{\ddagger\ddagger}}
\setfnsymbol{lamportnostar}

\newenvironment{sciabstract}{%
\begin{quote} \bf}
{\end{quote}}

\title{simPLE: a visuotactile method learned in simulation to precisely pick, localize, regrasp, and place objects}

\author
{Maria Bauza,$^{1\ast}$, Antonia Bronars,$^{1\ast}$ Yifan Hou,$^{2}$\footnote{Y. Hou did this work outside of Amazon. During the realization of this work, he was affiliated with the Robotics Institute at Carnegie Mellon University.} Ian Taylor,$^{1}$ \\
Nikhil Chavan-Dafle,$^{3}$ Alberto Rodriguez,$^{1}$\\
\\
\normalsize{$^{1}$Department of Mechanical Engineering, Massachusetts Institute of Technology,} \\ \normalsize{Cambridge, MA, 02139, USA}\\
\normalsize{$^{2}$ Amazon Robotics, Cambridge, MA, 02142, USA}\\
\normalsize{$^{3}$ Samsung AI Center,New York, NY, 10010, USA}\\
\\
\normalsize{$^\ast$ To whom correspondence should be addressed; Email: bauza@mit.edu, bronars@mit.edu.}\\
}

\date{}

\begin{document} 


\baselineskip24pt


\maketitle


\begin{sciabstract}
  Existing robotic systems have a clear tension between generality and precision. Deployed solutions for robotic manipulation tend to fall into the paradigm of one robot solving a single task, lacking precise generalization, i.e., the ability to solve many tasks without compromising on precision. This paper explores solutions for precise and general pick-and-place. In precise pick-and-place, i.e. kitting, the robot transforms an unstructured arrangement of objects into an organized arrangement, which can facilitate further manipulation. We propose simPLE (simulation to Pick Localize and PLacE) as a solution to precise pick-and-place. simPLE learns to pick, regrasp and place objects precisely, given only the object CAD model and no prior experience. We develop three main components: task-aware grasping, visuotactile perception, and regrasp planning. Task-aware grasping computes affordances of grasps that are stable, observable, and favorable to placing. The visuotactile perception model relies on matching real observations against a set of simulated ones through supervised learning. Finally, we compute the desired robot motion by solving a shortest path problem on a graph of hand-to-hand regrasps. On a dual-arm robot equipped with visuotactile sensing, we demonstrate  pick-and-place of 15 diverse objects with simPLE. The objects span a wide range of shapes and simPLE achieves successful placements into structured arrangements with 1mm clearance over 90\% of the time for 6 objects, and over 80\% of the time for 11 objects. Videos are available at \url{http://mcube.mit.edu/research/simPLE.html}.
\end{sciabstract}

\section*{Summary}
Robot precisely pick-and-places objects with hand-to-hand regrasps, visuotactile perception, and task-aware planning, given only the object CAD model and no prior experience.

% In setting up this template for *Science* papers, we've used both
% the \section* command and the \paragraph* command for topical
% divisions.  Which you use will of course depend on the type of paper
% you're writing.  Review Articles tend to have displayed headings, for
% which \section* is more appropriate; Research Articles, when they have
% formal topical divisions at all, tend to signal them with bold text
% that runs into the paragraph, for which \paragraph* is the right
% choice.  Either way, use the asterisk (*) modifier, as shown, to

\section*{Introduction}
Most deployed solutions for robotic manipulation fall into the paradigm of one-robot one-job, like repetitively welding from Point A to Point B. This limits their deployment into unstructured environments like homes, and hinders the automation industry from frequently and seamlessly  adapting and improving manufacturing lines. For effective deployment, adaptability is essential but not sufficient. Most jobs also depend on high accuracy and reliability. We suggest then that progress in robotic manipulation deployment truly relies on achieving precise generalization, i.e., solving many tasks without compromising on precision. This direction brings robotic manipulation closer to the paradigm of one-robot quickly adapting to and solving many-jobs.

\begin{figure*}[!ht]
\centering

\includegraphics[width=0.9\linewidth]{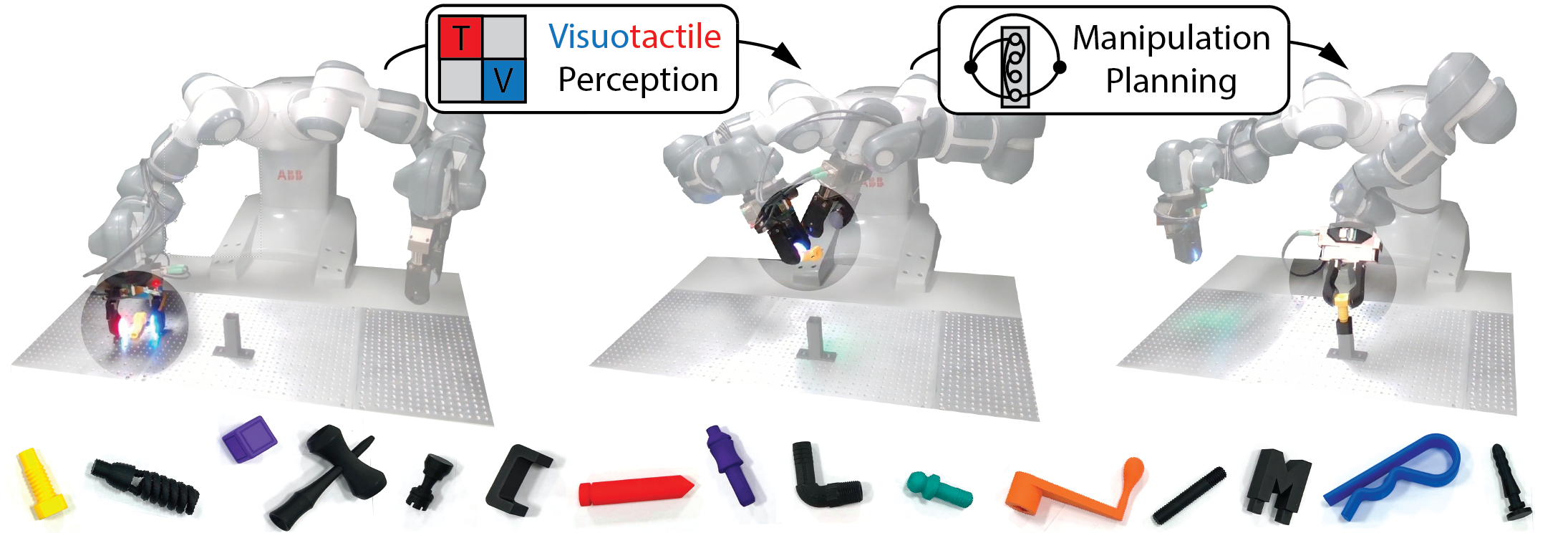}
\centering
\caption{\textbf{Precise pick-and-place with simPLE.} We present a system capable of precisely pick-and-placing objects learned entirely in simulation. The proposed solution consists of three models: task-aware grasping, visuotactile perception and motion planning. We show high fidelity transfer of the models to the real system for the 15 objects shown at the bottom of the figure.}\label{fig1}

\end{figure*}

Beyond being a longstanding goal in the automation industry, the task of precise pick-and-place can be a precursor for precise generalization. Precise pick-and-place requires transforming an unstructured pile of objects into a structured arrangement with objects placed in known locations, often with tight tolerances. It is a challenging task because the robot needs to pick up objects that it has never interacted with before and place them accurately in target configurations. In this work, we propose to solve precise pick-and-place with simPLE (simulation to Pick Localize and PLacE), an approach that precisely picks, regrasps and places objects by learning in simulation, given only the object CAD model and no prior experience.

\begin{figure*}[!ht]
\centering

\includegraphics[width=0.9\linewidth]{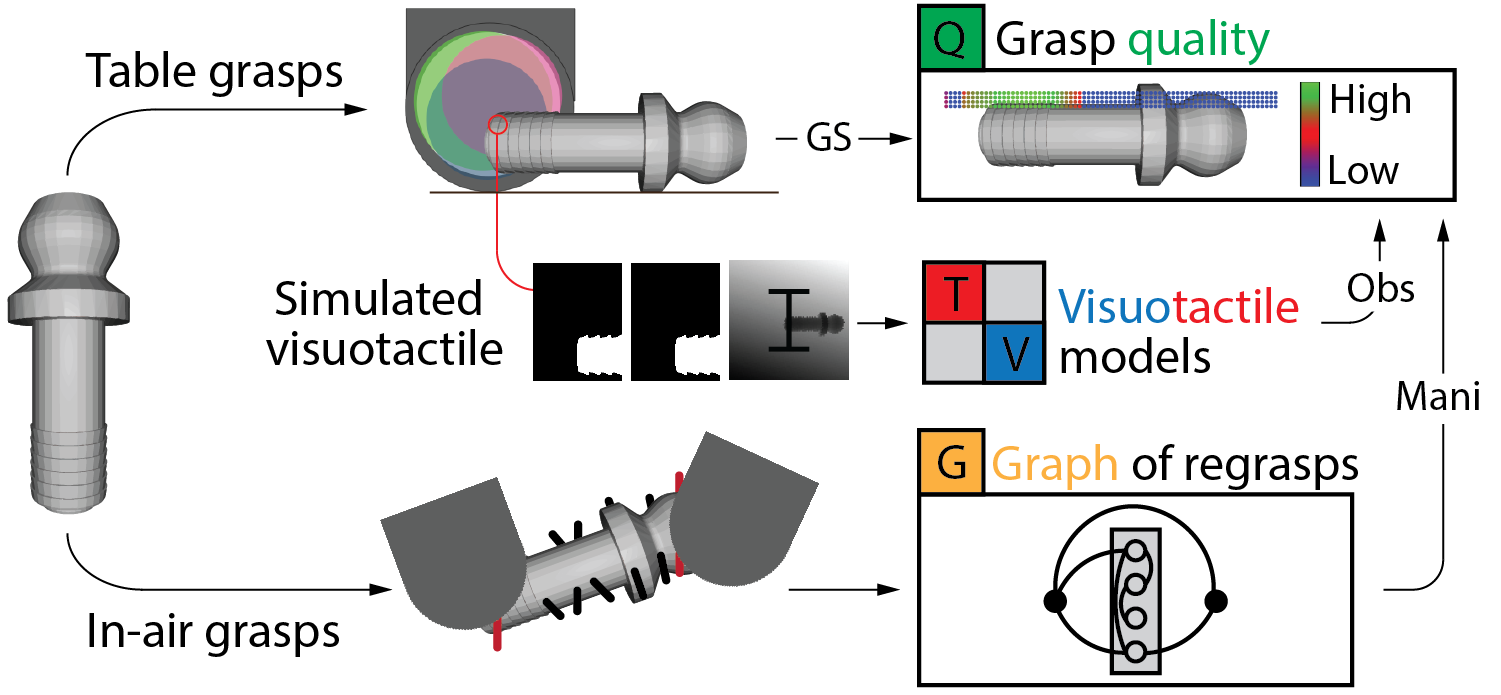}
\centering
\caption{\textbf{Generating models in simulation.} Starting from the object CAD model \textbf{(Left)}, we sample two types of grasps on the object. Table grasps \textbf{(Top)} are accessible from the object’s resting pose on the table. For each table grasp, we simulate corresponding depth and tactile images, and use these images to learn visuotactile perception models \textbf{(Center Right)}. In-air grasps \textbf{(Bottom)} are accessible during regrasps. We connect in-air grasp samples that are kinematically feasible into a graph of regrasps \textbf{(Bottom Right)}. We use the visuotactile model and regrasp graph to compute the observability (Obs) and manipulability (Mani) of a grasp, and combine these with grasp stability (GS) to evaluate the quality of each table grasp \textbf{(Top Right)}.}\label{fig2}

\end{figure*}

The simPLE solution to precise pick-and-place of novel objects is based on three components: task-aware picking, visuotactile object pose estimation, and motion planning using hand-to-hand regrasps, as depicted in Figure~\ref{fig1}. We design the components purely in simulation (Figure~\ref{fig2}) such that simPLE transfers to the real system without requiring any real prior experience with the objects (Figure~\ref{fig3}). We demonstrate simPLE in the context of a dual-arm robot equipped with tactile sensors and an external depth camera. From a depth image of the scene in front of the robot, we sample grasps on the object and estimate its pose. We score each grasp using a task-aware metric that accounts for the expected success of grasping, visuo-tactile object pose estimation, and pick-and-place motion planning; and then execute the grasp with the highest score. Next, our visuotactile approach updates the object pose by combining the estimated pose distributions from vision (before the grasp) and tactile (after the grasp). Finally, given the object pose we execute a motion plan for placing the object that requires solving a shortest path problem on a graph of hand-to-hand regrasps.

Existing work on grasping, the first challenge of precise pick-and-place, has shown clear progress toward generalization by being able to grasp many different types of objects from depth images alone \cite{zeng2022robotic,mahler2019learning,james2019sim}. Other works have explicitly estimated object shape or pose to obtain higher quality grasps \cite{chen2018probabilistic,deng2020self,chavan2022simultaneous}. While these grasping paradigms optimize for grasp stability, they do not consider the grasp’s usefulness to solve a downstream task (task-awareness). This is a limitation for full manipulation pipelines, where grasps should be functional as well as stable.

Current systems that demonstrate task-aware grasping often rely on large hand-annotated datasets of task-relevant grasps \cite{yang2019task,manuelli2019kpam,murali2021same,sun2021gater}, which is limited by the amount and quality of annotations. Other works rely on simulation to build task-aware metrics, but are limited to narrow task definitions \cite{lou2021collision,xu2022efficient}, or only show that the selected grasp can be achieved without attempting the task \cite{wen2022catgrasp}. \cite{fang2020learning} focuses on learning category-level task-oriented grasps for sweeping and hammering in simulation, without aiming at achieving high precision. \cite{he2023pick2place} use simulation for training placement-aware grasping; without considering the challenges of precision placement or the uncertainty in post-grasp object pose. By comparison, we learn task-aware metrics (graspability, observability and manipulability) in simulation and experimentally demonstrate that the chosen grasps facilitate precise placements.

In works that go beyond grasping, we find a division between achieving wide generalization and performing precise manipulations. In pick-and-place settings, approaches that consider objects with unknown geometries include learning from real experience \cite{berscheid2020self}, human annotations \cite{chen2022category}, or simulated data \cite{gualtieri2018learning,gualtieri2018pick,gualtieri2021robotic}. However these solutions tend to have wide tolerances for placements by targeting broad generality at the expense of precision. In this work, we build a system for precise pick-and-place that handles object generality by extending to any object with known geometry.

Other works that center on precise pick-and-place rely on hardware adaptations or simple object shapes, as well as known object geometry. For instance, \cite{morgan2021vision} achieves precise insertions by using actuator compliance, while \cite{kleeberger2021precise} combines suction and grasping to improve performance. \cite{kleeberger2020transferring}, which is most similar to our work, performs 6D grasps to increase precision, but lacks any regrasping strategy. This limits its applicability to objects and placements where there is a suitable grasp exposed. simPLE combines visuotactile information with task-oriented grasping to provide a solution that can handle any object and placement position, and is general by only requiring the object’s geometry.

One of the key features of our approach is relying on both visual and tactile information to achieve accurate perception. Without tactile, post-grasp displacements are hard to estimate \cite{zhao2021toward} and can impede precision. \cite{zhao2021towards} achieves precise placements through grasp selection, avoiding the problem of post-grasp displacements by relying on ground-truth perception after the grasp.

With tactile sensors, robots get direct access to contacts. However, the type of contact information depends on the sensor used.  \cite{nee2015task} finds that point tactile sensors can improve grasp stability, but carry limited information to estimate the object pose. Recent high-resolution sensors that rely on camera-based solutions \cite{yuan2017gelsight,lepora2021soft,lambeta2020digit,taylor2022gelslim} can facilitate accurate pose estimation \cite{bauza2022tac2pose,sodhi2021learning} or guide manipulation of simple geometries like boxes \cite{hogan2020tactile}, cables \cite{she2021cable} or connectors \cite{okumura2022tactile}. simPLE makes use of camera-based tactile sensors to combine visual and tactile information for accurate perception.

In summary, this work proposes an approach to manipulation that achieves generality by only requiring known object shapes rather than expensive real robot experience with those objects, and does not sacrifice precision by developing simulation tools that transfer well into real setups. We experimentally demonstrate successful pick-and-place of 15 diverse objects (Table~\ref{table:table1}), and through comparisons with baseline methods and ablation studies, we validate the need for visuotactile perception as well as task-aware planning (Table~\ref{table:table2}). Our approach makes an important step towards enabling more flexible solutions for general-purpose robotic pick-and-place.

\section*{Results}

\begin{figure*}[hp]
\centering

\includegraphics[width=0.45\linewidth]{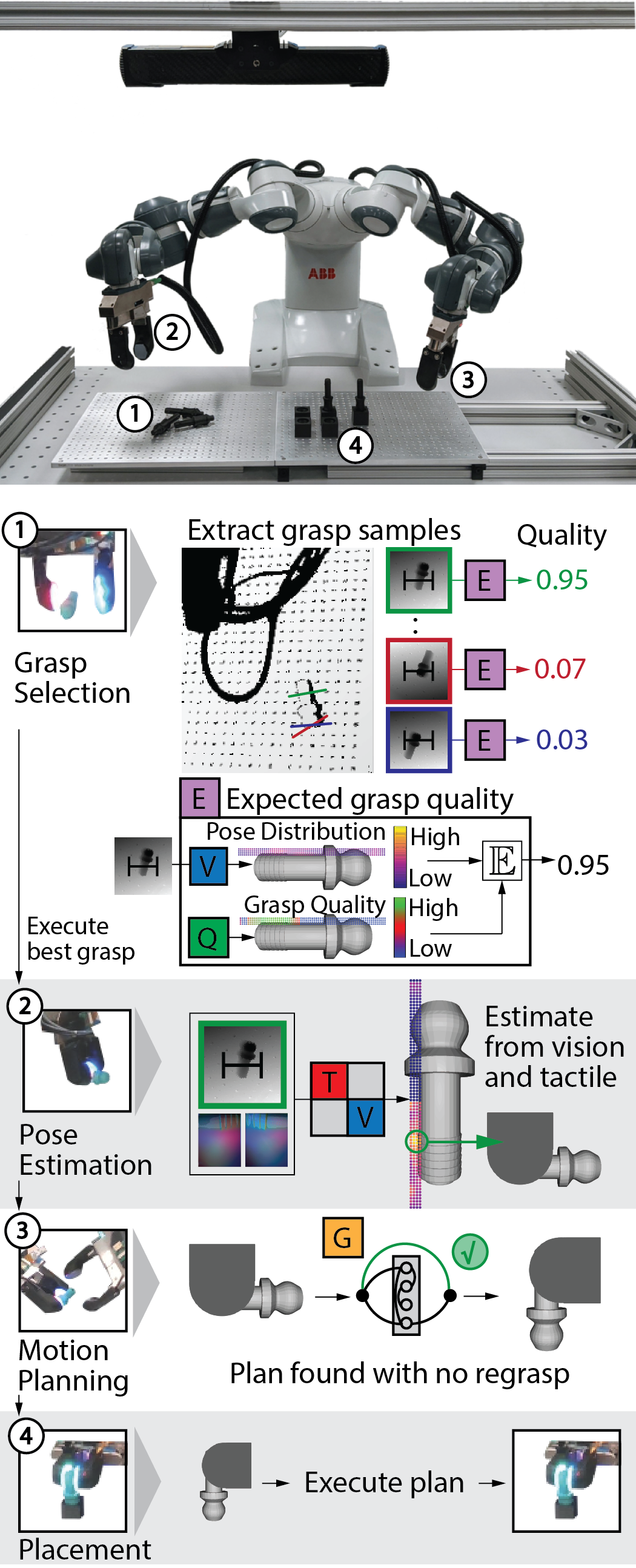}
\centering
\caption{\textbf{Deployment of simPLE.} simPLE first selects the best grasp from a set of samples on a depth image \textbf{(1)}. The best grasp is that which has the highest expected quality, given the pose distribution estimate from vision, and the precomputed grasp quality scores. Then, we execute the best grasp and update the pose estimate, now including information from tactile in addition to the original depth image \textbf{(2)}. Next, we take the best estimate from vision and tactile as the start pose, and find a plan that leads to the goal pose, using the regrasp graph if necessary \textbf{(3)}. Finally, we execute the plan \textbf{(4)}.}\label{fig3}
\end{figure*} 

This work aims at solving the task of precise pick-and-place purely in simulation so that robots can handle a large variety of objects without requiring direct experience. We split this task into three different steps:

\begin{enumerate}
    \item \textbf{Task-aware grasping.} From a depth image taken of the scene, simPLE samples antipodal grasps and computes an initial estimation of the object pose. Next, we assess the quality of each sampled grasp using a task-aware quality metric learned in simulation and command the robot to execute the best antipodal grasp (Figure~\ref{fig3}.1). 
    \item \textbf{Visuotactile object pose estimation.} Once an object is grasped, the robot receives as input tactile observations. Combining the tactile images with the initial depth image we update our estimate of the distribution of possible grasped object poses (Figure~\ref{fig3}.2). 
    \item \textbf{Motion planning.} Given the best estimate of the object pose, the robot computes the set of motions, including object regrasps if needed, that allow it to place the object at a desired configuration (Figure~\ref{fig3}.3). Finally, the robot executes the motions in open loop (Figure~\ref{fig3}.4).

\end{enumerate}

\subsection*{Learning robot models purely in simulation.}

To enable generality, simPLE learns the robot models in simulation without requiring prior experience. The learned models use shape information to assess the quality of grasps, estimate an object’s pose, and compute effective motion plans for placing.

Given an object’s CAD model, first, we sample a set of grasps on the object that are accessible from the object’s resting pose on the table (Figure~\ref{fig2}, Top). We denote these grasps as table grasps. Next, we render simulated versions of the visuotactile data we expect to observe for each table grasp. This consists of a pair of contact images (binary masks over the region of contact on each tactile sensor), and a depth image. The set of table grasps and their corresponding visuotactile data serves as a library of grasps that later we can match against real observations. For each table grasp, we also evaluate and store its task-aware quality, which corresponds to a composition of three metrics: graspability, observability, and manipulability.

The first metric, graspability, ranks different grasps depending on their capability to hold the object under disturbance forces. Using the simulated contact images, we can build a measure in simulation of graspability. Intuitively, the larger the contact region, the more force and torque the grasp can resist before the contact breaks or slips.

The second metric, observability, measures how likely is a grasp to produce tactile observations that facilitate the estimation of the object pose within the grasp. Computing the observability of a grasp requires having a model for pose estimation. Therefore, we start by learning in simulation how to estimate an object's pose by building a visuotactile perception model tailored to that object (Figure~\ref{fig2}, Center). These models learn to match visual and tactile observations to the simulated set of visuotactile data, outputting a distribution over object poses \cite{bauza2022tac2pose}. 

After learning how to perceive the object, we leverage the tactile model to evaluate the observability of each table grasp in simulation. Observability aims at quantifying how informative a given contact is in uniquely determining the object pose. Because tactile sensors provide a local view of the object geometry, many contacts may look similar and therefore provide ambiguous information about the object pose. Intuitively, any grasp that looks similar to a grasp on a distant region of the object is not a good indicator of object pose, and has low observability. Aiming for grasps with high observability allows a policy to prefer grasps with more unique features that, in turn, ease perception.

Finally, we want to compute the manipulability of each table grasp. Manipulability measures the simplicity of the best plan we can obtain from an initial table grasp, which includes the length of motion as well as the number of regrasps, i.e.  handing off the object to the other arm a number of times. Once an object is grasped at a known pose, placing it at a given configuration requires computing a motion plan that might include handing off the object to the other arm a number of times, i.e. regrasping the object. A regrasp may be necessary to increase the workspace of the robot, or if the initial grasp prevents collision-free placement of the object.

To compute this metric, we need to first develop a solution for finding motion plans. We start by obtaining in simulation a large set of in-air grasps (which is made of stable antipodal grasps). Then we find possible regrasps by checking which pairs are feasible kinematically, i.e., by computing if two grippers at each grasp in the pair would result in a collision-free situation. Precomputing the set of possible regrasps allows us to more efficiently solve the problem of finding online a feasible placing strategy.

Finding the best motion plan for placing an object now consists of building a graph of regrasps \cite{wan2015improving,hou2018fast} (Figure~\ref{fig2}, Bottom) and solving for the shortest path within it. For each table grasp, we use its grasp configuration as the initial node, and the desired placing pose as the goal node. Then solving the shortest path consists of finding if it is possible to directly place the object from its initial configuration (leading to a solution with no regrasp) or if applying one or more regrasps will then allow the robot to place the object. Edges in the regrasp graph represent kinematic feasibility between two nodes (aka if a pair of grasps is feasible or if it is possible to place the object from a given grasp without any collision). 

Manipulability penalizes the number of required regrasps in the shortest path from a given table grasp to the goal location. Table grasps that result in plans with fewer regrasps have higher manipulability scores, and are preferable.

Once we can compute the three metrics in simulation, we obtain and save the composite grasp quality for each table grasp as the product of its graspability, observability, and manipulability (Figure~\ref{fig2}, Top).

In summary, simulation allows us to compute the robot’s models required to predict pose distributions using visuotactile observations, compute motion plans to place the object in a given configuration, and assess the quality of table grasps using metrics for graspability, observability, and manipulability.

\subsection*{Experimental evaluation of simPLE in pick-and-place tasks}

We validate our approach on the task of precise pick-and-place, where rigid items of different shapes need to be picked up and placed precisely on rigid fixtures. This task represents a typical application in mechanical assembly and packing. Our robotic system consists of a dual arm robot with parallel hands, where tactile sensors are installed on the fingers. A top-down depth camera provides table top perception of the items.

Given a depth image, we follow the approach in \cite{mahler2019learning} to find possible antipodal grasps on the image. We then filter the grasp samples to avoid collisions between the robot’s fingers, the object, and the environment. For each collision-free grasp sample, we take a crop of the original depth image that is centered and aligned at the grasp (see Figure~\ref{fig3}) and provide it as input to the visual model which estimates the object pose as a distribution over table grasps. 

For each sampled grasp, we can compute its expected quality as the expectation of the quality metrics over the pose distribution provided by the visual model (Figure~\ref{fig3}.1). The grasp sample with the highest expected quality is the most likely to provide the best combination of graspability, observability, and manipulability.

If the expected best grasp is executed successfully, the robot will also receive tactile observations. Combining these with the previous depth image allows us to use the precomputed visuotactile models to update the vision-only estimate for the object pose (Figure~\ref{fig3}.2). Finally, we take the best pose estimate to compute the shortest path and find the simplest motion plan that the robot can execute to place the object (Figure~\ref{fig3}.3 and Figure~\ref{fig3}.4).

For each object, we conduct 20 trials in which the robot grasps the object from an unknown starting pose and attempts to place it in a known pose. Succeeding at a pick-and-place experiment requires placing the object into a tight cavity (Figure~\ref{fig1} and Figure~\ref{fig3}). We categorize each trial into success, near success, and failure. The experiments labeled as near success are cases where the object almost reached the goal position, and failure happened because of a small misalignment (of a few millimeters) during the final placement step, rather than an incorrect localization or a failed regrasp. Near successes could, in principle, be resolved by a closed-loop local insertion strategy.
We evaluate simPLE on 15 objects (Figure~\ref{fig1}). For 5 of the 15 objects, we also conduct a set of baseline experiments to evaluate the influence of task-aware grasping, tactile localization, and visual localization. Each baseline eliminates one of these components but preserves the other two. 

Tactile localization performs well in combination with task-aware grasping (tactile baseline) for objects where unambiguous grasp locations exist.  The success rate for grease, for example, is 19/20, and for rod, the tactile baseline achieves 12/20 successes and 5/20 near successes (Table~\ref{table:table2}). Some objects, such as hose, grip, and kendama, do not have any region where the grasp is entirely unambiguous. Therefore, even in the presence of task-aware grasping, tactile localization alone is unable to consistently resolve the object pose. This is reflected by a lower rate of successes and near successes for the tactile baseline for these objects (Table~\ref{table:table2}).

Visual localization performs well in combination with task-aware grasping (vision baseline) when the object configuration is unambiguous from an overhead camera. This is the case for hose, grip, kendama, and grease whose rates of successes and near successes are correspondingly high (Table~\ref{table:table2}). The object rod, on the other hand, has one threaded end and one unthreaded end, which is difficult to detect using only an overhead camera. This causes the vision baseline to fail 5/20 times. It is also noteworthy that the rate of near successes for hose and kendama are high relative to the rate of true successes. This is because while vision is capable of globally disambiguating between object orientations, it is less adept at refining the object pose to high precision, which is necessary to successfully place the object into the cavity. When tactile localization is included, we see more true successes.

\begin{figure*}[ht]
\centering

\includegraphics[width=\linewidth]{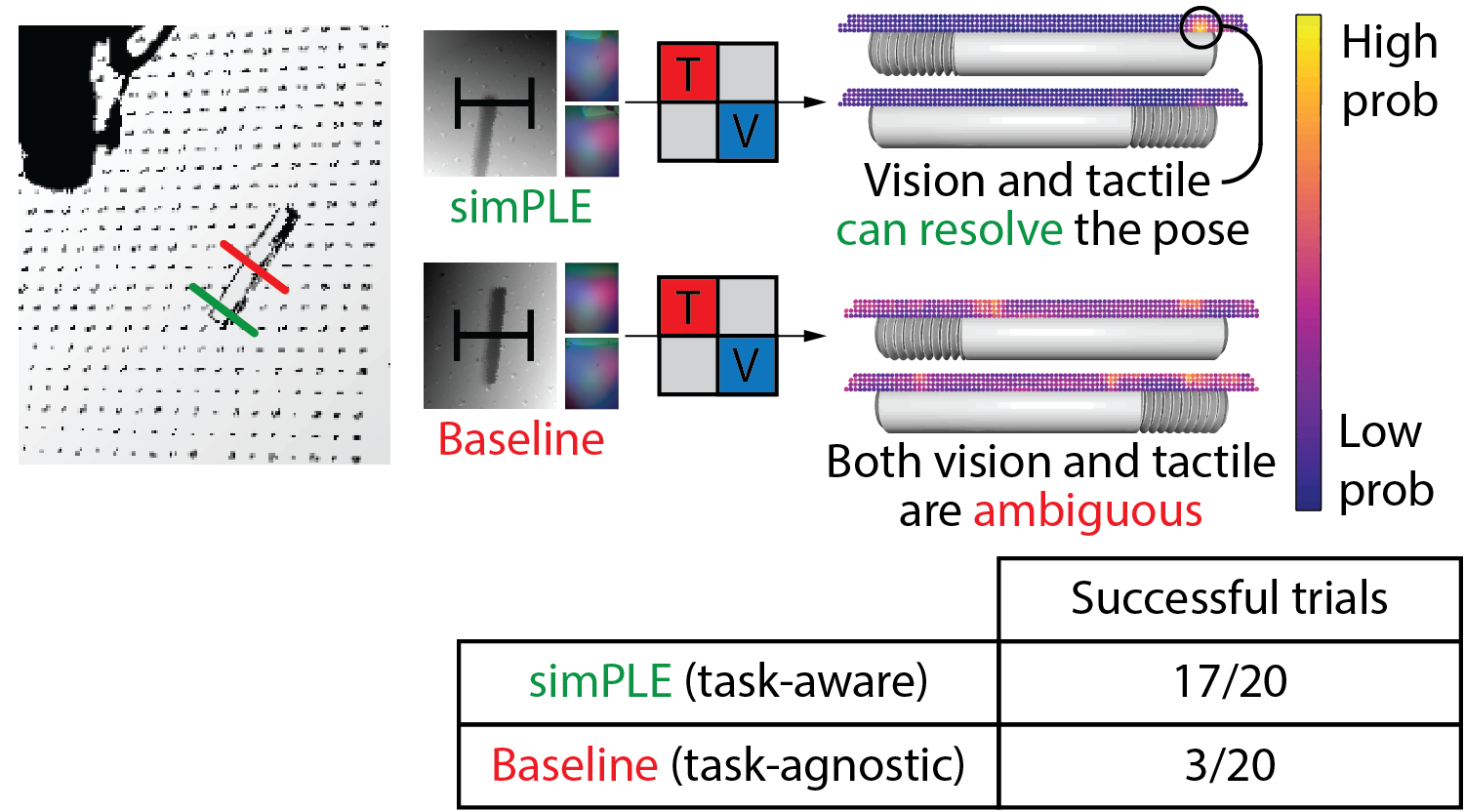}
\centering
\caption{\textbf{Task-aware grasping facilitates perception.} We consider the case of the object rod and compare simPLE against a baseline that doesn’t perform task-aware grasping. We show the type of grasp selected by each method and the final pose distribution after grasping the object. The baseline cannot resolve the pose, as both tactile and vision observations are ambiguous. Instead, simPLE aims at grasps that it expects will produce observable tactile imprints, allowing the perception model to resolve the pose after the grasp. As a result, simPLE succeeds in 17 out of 20 real experiment trials, while the baseline, due to perception errors, only succeeds 3 times (\textbf{Bottom}). 
}\label{fig4}

\end{figure*}

Next, we compare simPLE against a task-agnostic baseline which uses the quality metric from Dexnet \cite{mahler2019learning} and therefore is task agnostic and object-independent. Note that simPLE instead takes into account the downstream goals of localization and placement, in addition to object-tailored graspability, before choosing a grasp. Instead, the task-agnostic baseline considers only the likelihood of grasp success when choosing a grasp. For some objects, like hose and kendama, the task-agnostic baseline often prefers the same grasp as simPLE, and therefore the success rates are similar. For other objects, like rod and grease, task awareness leads simPLE to prefer a different grasp than the task-agnostic baseline.

For the object rod, we find that task-aware grasping facilitates perception, by targeting grasps that lead to unambiguous tactile information (Figure~\ref{fig4}). This is particularly important for rod because visual information alone is ambiguous. simPLE results in 17/20 successful trials, compared with 3/20 successful trials for the task-agnostic baseline. The task-agnostic baseline targets grasps near the center of the object, where both vision and tactile information are ambiguous. simPLE is able to resolve the pose by targeting grasps near the end of the object, where tactile information can resolve the ambiguity in the visual information.

The object rod also highlights an important benefit of our regrasp-based planning framework - it allows for recovery from vision-based perception failures. In six of simPLE’s 17 successful trials, the robot grasps on the wrong end of rod because of a vision ambiguity (the two ends of the object are difficult to disambiguate from vision alone). After incorporating tactile information, the robot amends its estimate, and plans a regrasp in order to place the object in the correct orientation. The simPLE preferred grasp for this object is expected to result in a placement without a regrasp. However, the vision confusion and the consequent correction from tactile information, triggers a correction regrasp by the motion planning framework that allows the robot to recover from this vision failure.

\begin{figure*}[!ht]
\centering

\includegraphics[width=\linewidth]{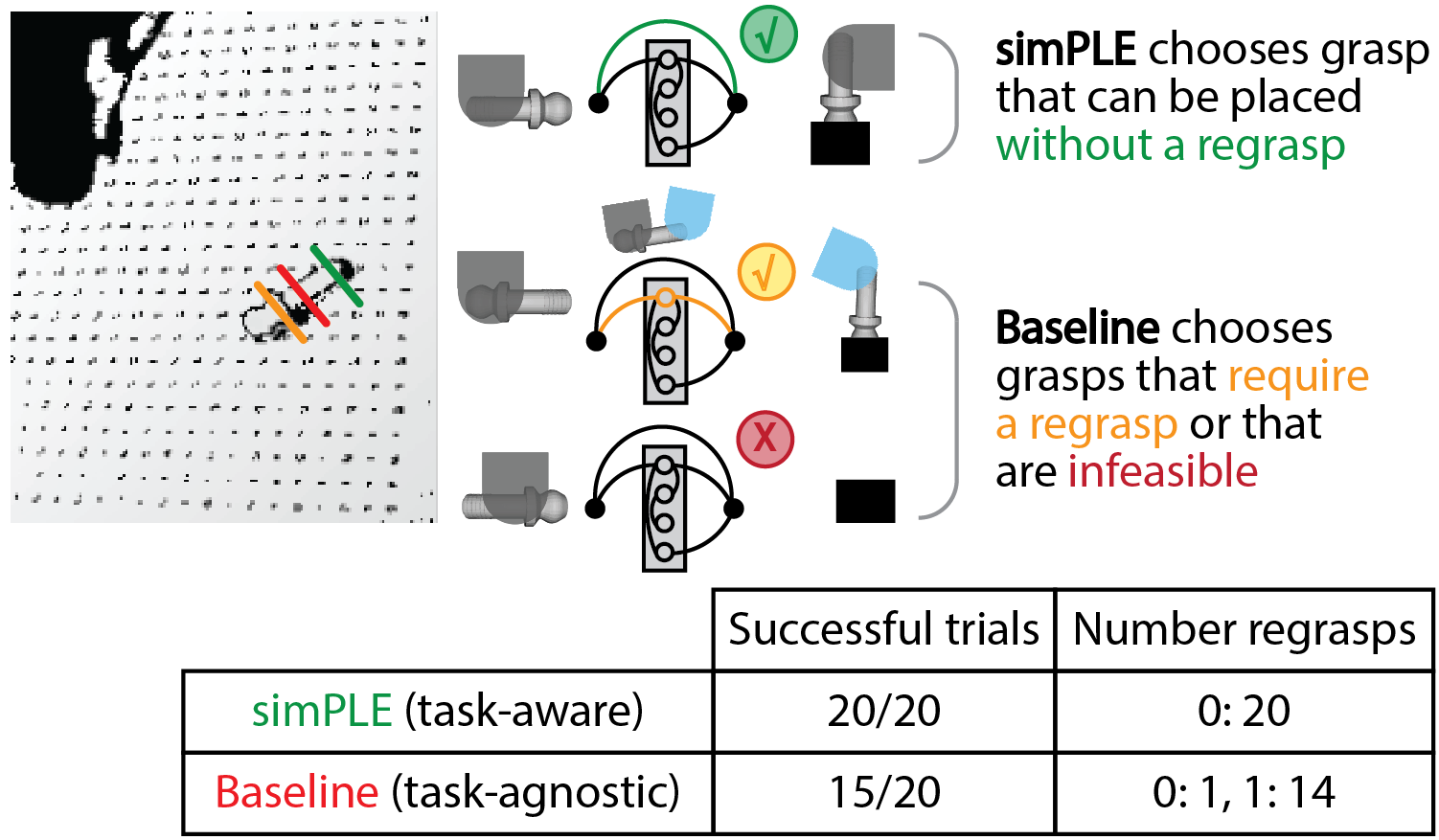}
\centering
\caption{\textbf{Task-aware grasping results in better planning solutions.} A baseline that doesn’t aim at task-aware grasping can end up choosing grasps that require more regrasps than needed, or that result in an object configuration where no motion plan is possible. Instead, simPLE chooses grasps that reduce the number of regrasps required to place the object. Aiming at grasps that require simpler planning strategies ends in more successful trials (\textbf{Bottom}).}\label{fig5}

\end{figure*} 

As another example, for the object grease, we find that task-aware grasping facilitates motion planning by targeting grasps that lead to successful motion plans (Figure~\ref{fig5}). Because grease is a small object, regrasps are infeasible from some table grasps. This can make it impossible to place the object if motion planning is not considered before the first grasp. simPLE results in 20/20 successful trials, compared with 15/20 successful trials for the task-agnostic baseline. The task-agnostic baseline aims at grasps near the center of the object that require a regrasp. In the 5/20 failed trials, the initial grasp does not leave enough room for a regrasp, and thus the attempt fails. The remaining 15 trials of the task-agnostic baseline are successful, but they require a regrasp that simPLE avoids. simPLE targets grasps that require simpler planning strategies and result in a higher success rate.

\begin{figure*}[!ht]
\centering

\includegraphics[width=\linewidth]{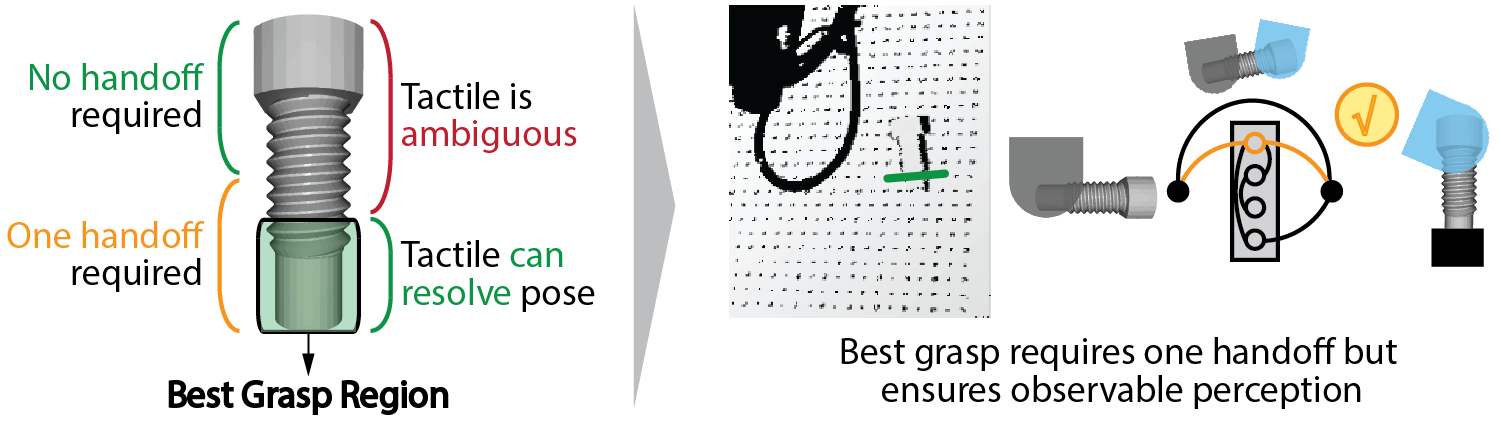}
\centering
\caption{\textbf{Task-aware grasping balances perception and planning to find the grasp region that maximizes success.} For the object stud, there is no one grasp region that maximizes both observability and manipulability (\textbf{Left}). simPLE chooses a grasp that requires a hand-off in order to ensure observable perception (\textbf{Right}), negotiating the trade-off between observability and manipulability in order to maximize task success.
.}\label{fig6}

\end{figure*} 

We test simPLE on a total of 15 objects to evaluate the method’s generalization to a wide variety of object shapes and sizes. Table~\ref{table:table1} shows the number of attempts that succeeded after 20 trials, as well as the number of near successes and failures. Overall, simPLE provides the robot with a method that successfully transfers to the real world and achieves precise pick-and-place. For 9 of the objects, simPLE gets a success rate of at least 85\%. Only magnet has a success rate of less than 50\%. The object magnet is particularly challenging because both visual and tactile information are not sufficiently unique in many of its orientations.

\begin{table*}[ht]
    \caption{\textbf{simPLE success rate on 15 objects.} The rate of success, near success, and failure out of 20 trails for 15 objects is tabulated in order of highest success rate, to lowest.}
    \label{table:table1}
    \begin{center}
        \begin{tabular}{|c|c|c c c|}
             \hline
             \multicolumn{2}{|c|}{\textbf{Object}} & \makecell{\textbf{Successes} \\ \textbackslash 20 trials} & \makecell{\textbf{Near Successes} \\  \textbackslash 20 trials} & \makecell{\textbf{Failures} \\ \textbackslash 20 trials}  \\ 
             \hline
             Grease &
             \makecell{
                \begin{minipage}{15mm}
                    \vspace{1pt}
                    \centering
                    \includegraphics[height=6mm, angle=0]{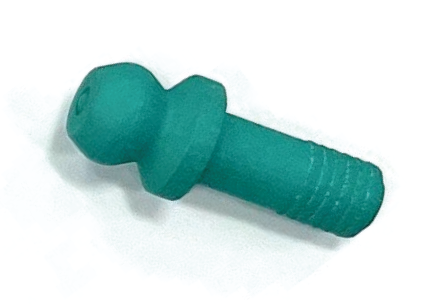}
                    \vspace{1pt}
                \end{minipage}} & 20 & 0 & 0 \\
             Head &
             \makecell{
                \begin{minipage}{15mm}
                    \vspace{1pt}
                    \centering
                    \includegraphics[height=6mm, angle=70]{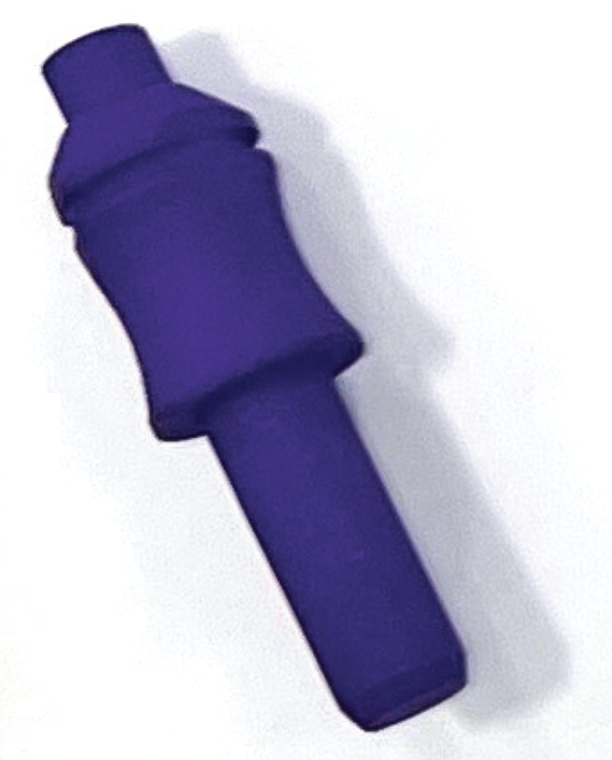}
                    \vspace{1pt}
                \end{minipage}} & 20 & 0 & 0 \\
             Cube &
             \makecell{
                \begin{minipage}{15mm}
                    \vspace{1pt}
                    \centering
                    \includegraphics[height=6mm, angle=10]{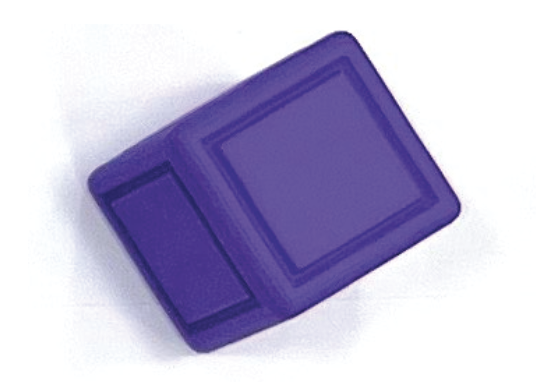}
                    \vspace{1pt}
                \end{minipage}} & 19 & 1 & 0 \\
             Cotter & 
             \makecell{
                \begin{minipage}{15mm}
                    \vspace{1pt}
                    \centering
                    \includegraphics[height=6mm, angle=0]{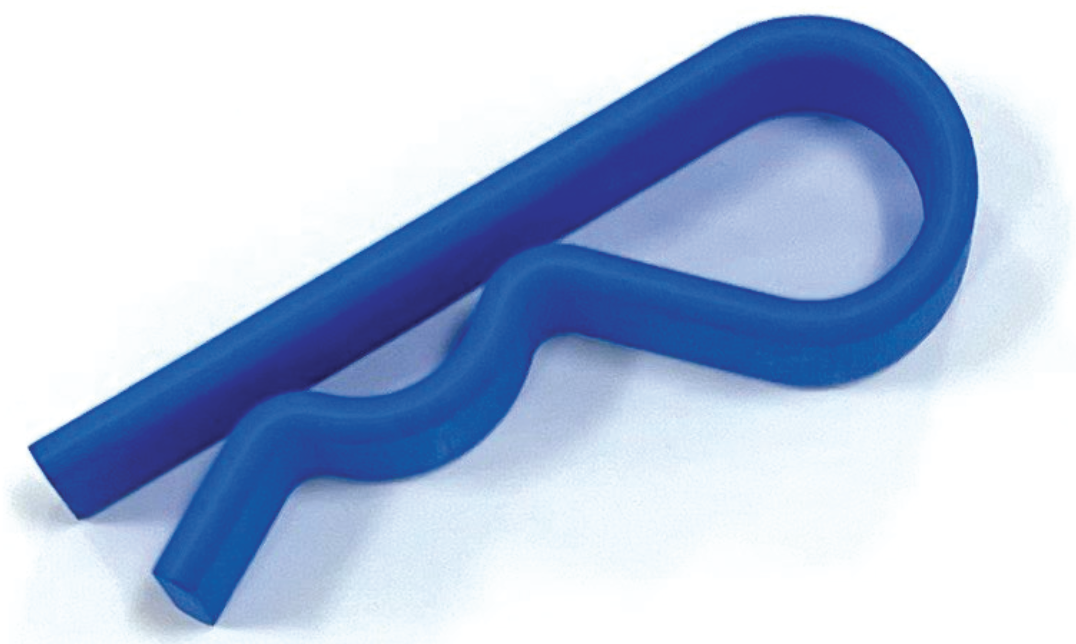}
                    \vspace{1pt}
                \end{minipage}} & 19 & 0 & 1 \\
             Pin & 
             \makecell{
                \begin{minipage}{15mm}
                    \vspace{1pt}
                    \centering
                    \includegraphics[height=9mm, angle=90]{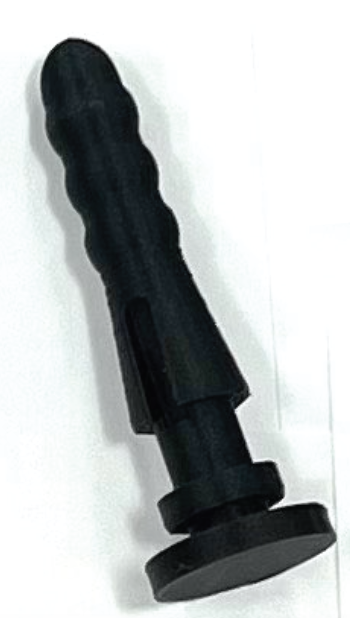}
                    \vspace{1pt}
                \end{minipage}} & 18 & 2 & 0 \\
             Pencil &  
             \makecell{
                \begin{minipage}{15mm}
                    \vspace{1pt}
                    \centering
                    \includegraphics[height=5mm, angle=0]{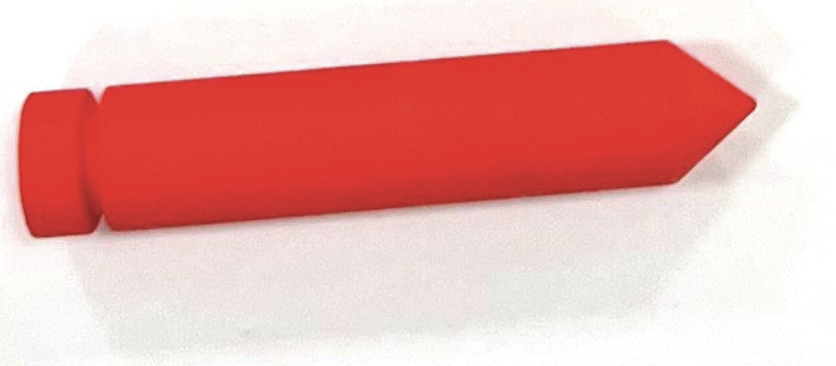}
                    \vspace{1pt}
                \end{minipage}} & 18 & 0 & 2 \\
             Grip &  
             \makecell{
                \begin{minipage}{15mm}
                    \vspace{1pt}
                    \centering
                    \includegraphics[height=6mm, angle=0]{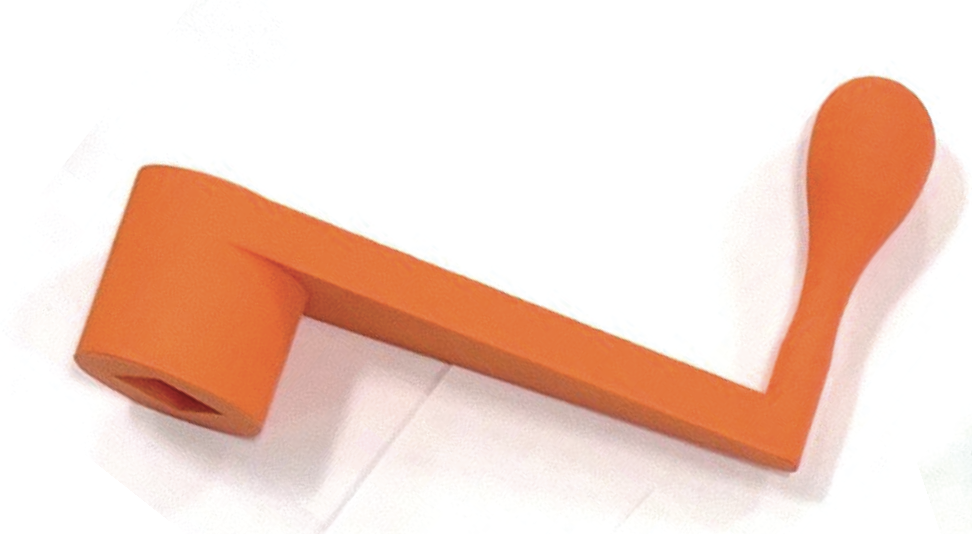}
                    \vspace{1pt}
                \end{minipage}} & 17 & 3 & 0 \\
             Letter &  
             \makecell{
                \begin{minipage}{15mm}
                    \vspace{1pt}
                    \centering
                    \includegraphics[height=6mm, angle=0]{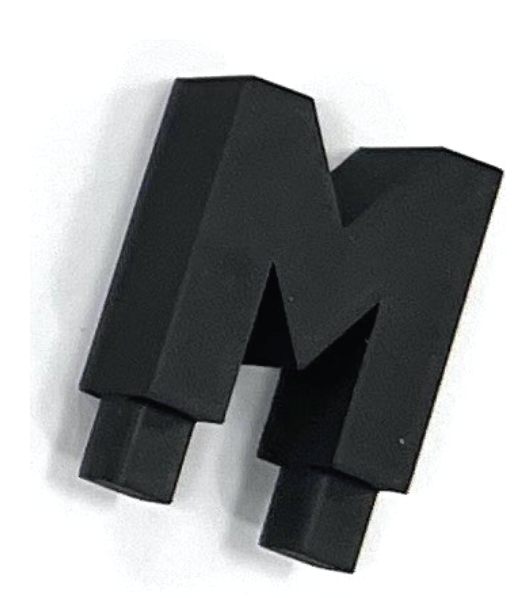}
                    \vspace{1pt}
                \end{minipage}} & 17 & 2 & 1 \\
             Rod &  
             \makecell{
                \begin{minipage}{15mm}
                    \vspace{1pt}
                    \centering
                    \includegraphics[height=6mm, angle=-20]{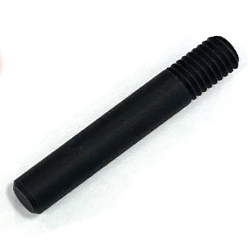}
                    \vspace{1pt}
                \end{minipage}} & 17 & 1 & 2 \\
             Stud &  
             \makecell{
                \begin{minipage}{15mm}
                    \vspace{1pt}
                    \centering
                    \includegraphics[height=10mm, angle=90]{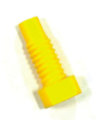}
                    \vspace{1pt}
                \end{minipage}} & 16 & 4 & 0 \\
             Kendama &  
             \makecell{
                \begin{minipage}{15mm}
                    \vspace{1pt}
                    \centering
                    \includegraphics[height=6mm, angle=-20]{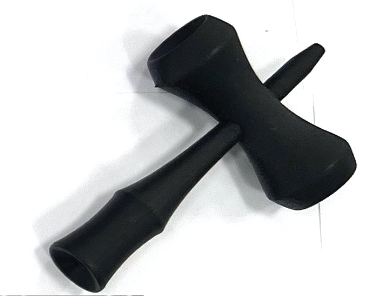}
                    \vspace{1pt}
                \end{minipage}} & 16 & 3 & 1 \\
             Rook &  
             \makecell{
                \begin{minipage}{15mm}
                    \vspace{1pt}
                    \centering
                    \includegraphics[height=8mm, angle=-90]{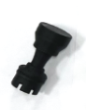}
                    \vspace{1pt}
                \end{minipage}} & 15 & 2 & 3 \\
             Hose &  
             \makecell{
                \begin{minipage}{15mm}
                    \vspace{1pt}
                    \centering
                    \includegraphics[height=6mm, angle=0]{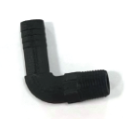}
                    \vspace{1pt}
                \end{minipage}} & 14 & 3 & 3 \\
             Lamp &  
             \makecell{
                \begin{minipage}{15mm}
                    \vspace{1pt}
                    \centering
                    \includegraphics[height=6mm, angle=20]{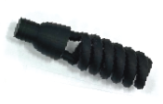}
                    \vspace{1pt}
                \end{minipage}} & 10 & 10 & 0 \\
             Magnet & 
             \makecell{
                \begin{minipage}{12mm}
                    \vspace{1pt}
                    \centering
                    \includegraphics[height=6mm, angle=-60]{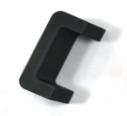}
                    \vspace{1pt}
                \end{minipage}} & 7 & 5 & 8 \\
             \hline
        \end{tabular}
    \end{center}
\end{table*}

\begin{table*}[ht]
    \caption{\textbf{simPLE success rate vs. baseline on 5 objects.} The rate of success, near success, and failure for 5 objects on simPLE and 3 baselines. simPLE and the 3 baselines were evaluated with 20 trails each.}
    \label{table:table2}
    \begin{center}
        \renewcommand{\arraystretch}{1.5} % Default value: 1
        \begin{tabular}{|c|c c c |c c c| c c c | c c c |}
        \hline
        \multirow{2}{*}{\textbf{Object Name}} & \multicolumn{3}{c|}{\textbf{simPLE}} & \multicolumn{3}{c|}{\textbf{Agnostic Baseline}} & \multicolumn{3}{c|}{\textbf{Tactile Baseline}} & \multicolumn{3}{c|}{\textbf{Vision Baseline}} \\
        \cline{2-13}
        & \makecell{\vspace{3pt} S\\\textbackslash 20 \vspace{3pt}} & \makecell{ \vspace{3pt} NS\\\textbackslash 20 \vspace{3pt}} & \makecell{\vspace{3pt} F\\\textbackslash 20 \vspace{3pt}} & \makecell{\vspace{3pt} S\\\textbackslash 20 \vspace{3pt}} & \makecell{\vspace{3pt} NS\\\textbackslash 20 \vspace{3pt}} & \makecell{\vspace{3pt} F\\\textbackslash 20 \vspace{3pt}} & \makecell{\vspace{3pt} S\\\textbackslash 20 \vspace{3pt}} & \makecell{\vspace{3pt} NS\\\textbackslash 20 \vspace{3pt} } & \makecell{ \vspace{3pt} F\\\textbackslash 20 \vspace{3pt}} & \makecell{ \vspace{3pt} S\\\textbackslash 20 \vspace{3pt}} & \makecell{ \vspace{3pt} NS\\\textbackslash 20 \vspace{3pt}} & \makecell{ \vspace{3pt} F\\\textbackslash 20 \vspace{3pt}} \\
        \hline
        Grease & \textbf{20} & 0 & 0 & \textbf{15} & 0 & 5 & \textbf{19} & 0 & 1 & \textbf{19} & 0 & 1 \\
        Grip & \textbf{17} & 3 & 0 & \textbf{10} & 2 & 8 & \textbf{1} & 0 & 19 & \textbf{18} & 2 & 0 \\
        Rod & \textbf{17} & 1 & 2 & \textbf{3} & 8 & 9 & \textbf{12} & 5 & 3 & \textbf{15} & 0 & 5 \\
        Kendama & \textbf{16} & 3 & 1 & \textbf{15} & 2 & 3 & \textbf{10} & 3 & 7 & \textbf{14} & 6 & 0 \\
        Hose & \textbf{14} & 3 & 3 & \textbf{15} & 3 & 2 & \textbf{8} & 2 & 10 & \textbf{12} & 6 & 2 \\
        \hline
        \end{tabular}
    \end{center}
\end{table*}

Out of the several objects studied, we highlight the case of stud as it provides an important discussion point for simPLE. As depicted in Figure~\ref{fig6}, stud’s best table grasps, according to the task-aware quality, require at least one regrasp if we want to ensure tactile observability. During experiments, simPLE selected these grasps and has a success rate of 80\% with 20\% near successes and no failures. Therefore all selected table grasps required a regrasp. Choosing other table grasps would not require a regrasp but that would come at the cost of losing tactile observability and thus a higher likelihood of failing to estimate the right object pose. Dealing with such trade-offs is inevitable as task awareness implies balancing multiple objectives (graspability, observability, and manipulability).

\section*{Discussion}

In this work, we present an approach to precise pick-and-place that leverages offline computing to allow a high degree of adaptability. Being able to use the same algorithms and system to precisely place new objects taken from unstructured scenes is at the core of many manipulation problems that still remain unsolved. By not requiring prior robot or human experience with those objects, we showcase that it is possible to aim at building flexible systems without compromising on accuracy. 

Our experiments suggest that through visuotactile perception and task awareness, robot models learned in simulation can successfully transfer to real systems. We show this for a large variety of objects sizes and shapes. When compared to baselines, our results validate the need for having both tactile and visual sensing, as well as the benefits of deploying policies that consider the task requirements  end-to-end.

\subsection*{Simulated observations to learn robot models}

While it is common to learn behavior policies purely in simulation, here we use simulation to learn precise perception models that can then inform model-based behavior policies. Learning perception purely in simulation allows us to generate the large amounts of data needed to predict meaningful pose distributions. By simulating observations, we are also able to rethink perception as a matching problem (how similar are observations?) rather than a regression (which pose could produce a given observation?). Matching is simpler and allows us to easily generate probabilities as we can quantify how similar (likely) different observations are.

Combining vision, which is global, with tactile, which is local but provides higher resolution at contact, makes our system capable of precisely estimating poses for objects with more variety in shapes, textures, and sizes.

\subsection*{Task awareness to select grasps}

Our experiments suggest that accounting only for the success of a grasp is less effective than selecting task-aware grasps that also account for facilitating perception and planning. We show this in more detail in Figure~\ref{fig4} and Figure~\ref{fig5} where selecting task-aware grasps is key to resolving the object pose and to finding placing plans with fewer regrasps, respectively. During experiments, we measure the task-aware quality of a grasp by taking the product of 3 metrics: graspability, observability, and manipulability. Nevertheless, simPLE would allow us to seamlessly integrate other relevant metrics like penalizing grasps on deformable or fragile regions of an object. To further improve our solution, we believe it would be useful to systematically assess the importance of each metric towards solving the task. This could lead to better ways of defining and combining metrics. For instance, ensuring good pose estimation might be of higher or lower importance than simplifying a motion plan for end-to-end reliability.

\subsection*{Scope of applicability}

simPLE can accommodate different sensor modalities as long as it is possible to simulate their observations. Following the process that simPLE uses for our tactile and vision sensors, we would learn a perception model based on matching real observations to a set of precomputed ones. Since each model outputs a distribution of likelihoods over possible poses, integrating the sensing modalities reduces to simply multiplying the likelihoods obtained from each of the observations \cite{bauza2022tac2pose}. 

Also relevant to the applicability of simPLE, it can integrate task constraints such as avoiding environment collisions or dealing with multiple objects on the scene. Qualitative experiments have shown that simPLE for grasp selection is able to find good grasps on the objects even when multiple objects (of different types) are present. In the cases where there are multiple types of objects, the robot can run grasp selection for each object type and select the grasp with highest quality which solves the object classification problem implicitly.

\subsection*{Opportunities for future work}

simPLE shows that it is possible to build adaptable solutions for robotic manipulation while achieving high accuracy. To that end, simPLE features an open-loop solution based on learning from simulated data. Next we expose some of its limitations and comment on how to overcome them:

\begin{itemize}
    \item \textbf{Near success failures.} As described in the results section, it is not uncommon to have near success due to small misalignments that end in the object being placed quite close yet outside the ±0.5mm placing tolerance. Such cases would benefit from closed-loop execution during the placement. For instance, we could measure the forces experienced during placing and detect external contacts \cite{kim2022active}. This could allow the robot to identify the evolution of the placement as well as derive a policy for correcting deviations.
    \item \textbf{Reacting to the unexpected.} Currently, simPLE is open-loop in that after receiving tactile images from the initial grasp and producing a visuotactile pose estimate, it doesn’t update its belief of the object pose. While we show that this is sufficient to get relatively good results, in some cases it proves insufficient. Adding a tracking strategy that aggregates multiple sensor readings over time would help avoid cases of ambiguity and reduce failure. A clear step in the system that would benefit from additional feedback is after regrasps, which tend to exacerbate error. Such a strategy could also help deal with unexpected disturbances like the object sliding in the grasp due to collisions with a dynamic or unknown environment. 
    \item \textbf{Defining observability under multiple sensors.} Currently, the notion of observability only takes into account how likely a grasp is to produce useful observations for tactile perception. This means that we are not accounting for vision which in some cases would be sufficient to estimate the pose even if tactile would be ambiguous. We believe that building the algorithmic and mathematical tools to better understand and exploit observability is paramount. Defining observability after aggregating multiple sensors would increase the range of poses where the object pose can be estimated. Moreover, it would also be possible to provide the robot with policies that take multiple actions, rather than a single grasp as in simPLE, to ensure higher observability and thus avoid wide or multimodal pose distributions.
\end{itemize}

Defining observability under multiple sensors. Currently, the notion of observability only takes into account how likely a grasp is to produce useful observations for tactile perception. This means that we are not accounting for vision which in some cases would be sufficient to estimate the pose even if tactile would be ambiguous. We believe that building the algorithmic and mathematical tools to better understand and exploit observability is paramount. Defining observability after aggregating multiple sensors would increase the range of poses where the object pose can be estimated. Moreover, it would also be possible to provide the robot with policies that take multiple actions, rather than a single grasp as in simPLE, to ensure higher observability and thus avoid wide or multimodal pose distributions.

\section*{Materials and Methods}

\subsection*{Notation}

Let $t$ denote a grasp that could occur without collision between the robot's fingers, a flat environment like a table, and an object resting on top of it. We denote $T$ as a discrete set taken from all the possible table grasps and $p(t = \argmin_{t' \in T} dist(t', t_o) \in T, o)$ as the density function that represents how likely a given grasp $t \in T$ is to be the closest in pose distance (measured by the function $dist$) to a table grasp $t_o$ that produced the observation $o$. To simplify notation we will denote the probability above as $p(t | o)$.  The dist function corresponds to the ADD \cite{tekin2018real} (average 3D distance) metric which measures the average distance between the pointcloud of the object in two different poses.

In our case, an observation $o$ of a table grasp $t$ includes a simulated visual observation from the depth camera in the form of a depth image, $d$, and a pair of simulated tactile images from the tactile sensors, $c$, i.e. $o=(d, c)$. Our perception models then estimate the following distribution: $p(t | d, c)$ which provides the likelihood of each table grasp in $T$ given the visuotactile observations $d$ and $c$ (which are simulated observations during learning, and real observations at test time).

For each table grasp, we develop functions to simulate several of its attributes such as the expected contacts $\hat{c}(t)$ on each finger of the grasp, simulated depth image $\hat{d}(t)$ of the grasp (see Figure~\ref{fig2}), and quality score, $q(t)$. Each table grasp $t$ is uniquely defined by considering the pose of the object w.r.t. to the gripper frame. We will refer to such pose also as $t$ to simplify notation.

We also consider in-air grasps, denoted as $g$, which correspond to grasps that could happen without collisions between the robot fingers and the object. Note that now the environment is not constraining the grasp and thus in-air grasps include but are not restricted to being table grasps. For in-air grasps, we consider the notion of regrasps, $r$, to denote whether two grippers grasping an object at the grasp locations $g_1$,$g_2$ would create any collisions. Therefore $r(g_1,g_2) = {0,1}$  where $0$ denotes the presence of some collision while $1$ indicates that the regrasp is collision-free. With this definition of regrasp, we can build a graph of regrasps, $\mathcal{G} = \mathcal{G}(t_{init},q_{goal})$, where its nodes include a discretized set of in-air grasps, $G$, as well as the initial table grasp, $t_{init}$, and the desired goal configuration of the object at placement, $q_{goal}$. The edge between grasps $g_1,g_2 \in G$ on the graph $\mathcal{G}$ are precomputed and exist if $r(g_1,g_2) = 1$. At runtime,  start ($t_{init}$) and goal ($q_{goal}$) nodes are added, as well as edges that connect them to the pre-computed graph.

\subsection*{Table grasps}

Table grasps are the set of grasps that could occur without collision between the robot fingers, a flat environment like a table, and an object resting on top of that table. We generate a discrete set $T$ of table grasps by first determining the resting configurations of the object on a flat table. Then, we sample antipodal grasps with 1mm resolution around the object. We constrain the height of the antipodal samples by enforcing that the tips of the robot fingers are flush with the table during the grasp. We augment the set of table grasps by perturbing the object from its resting configuration to account for any object perturbations that may occur during grasping. We consider small perturbations about the axis of the grasp, and of the height of the grasp. The set of table grasps $T$ consists of poses, $t$ , measured relative to the gripper and the corresponding set of simulated visuotactile data for those poses, $\hat{c}(t)$ and $\hat{d}(t)$. The process for rendering visuotactile data in simulation is described below.

\subsection*{Rendering observations in simulation}

We use a similar procedure to simulate contact and depth images to represent the actual observations from real sensors. How to simulate contact is extensively described in \cite{bauza2022tac2pose}  and consists of rendering images from a virtual camera of an object placed such that its closest point to the virtual camera would contact without penetrating an imaginary sensor. The virtual camera renders a depth image from the scene which we then process to create a contact mask using the pixels where the object would penetrate the sensor.

Simulating depth images requires rendering both the table and object at a given position and taking a depth image using a virtual camera that matches the extrinsic and intrinsic parameters of the real one. Then we can compute the depth image associated with each table grasp by taking a crop of the rendered depth image by centering and reorienting it at the pose of the table grasp \cite{mahler2019learning}. The resulting crop as well as the contact images simulated for each of the fingers on the gripper provide the set of simulated observations for each table grasp.

\subsection*{Visuotactile pose estimation}

For each object and sensor modality (vision and tactile in our case), we train an encoder to match observations to the pre-computed set of observations in the set of table grasps $T$, following the approach outlined in \cite{bauza2022tac2pose}. Once we have learned to encode observations based on their distances in pose space, we pre-compute encodings for the observations in the set $T$. At test time, we compute a new encoding for the real observation coming from the actual sensor and compare its distance in embedding space to all of the precomputed encodings in $T$. We apply a softmax to the resulting vector of distances to obtain a probability distribution over the table gasps in $T$. This distribution represents the likelihood of the object configuration matching each element in $T$, given the real observation.

For each table grasp $t$, we precompute encodings for its two contact images as well as the depth image that represents the expected observation at that grasp. This allows us to efficiently compute the distributions over possible object poses for each sensing modality online.

During experiments, visuotactile observations consist of a depth image, $d$, of the object before grasping, two contact images $c$, and the gripper width during a grasp. The final estimate of the pose distribution,   $p(t | d, c)$, comes from the product of the distribution obtained from the depth image  $p(t | d)$, the distributions obtained from the contact images   $p(t | c)$, and a Gaussian centered at the gripper width.

\subsubsection*{Training details}
The visual encoder and the tactile encoder are trained jointly. Each encoder consists of a ResNet-50 \cite{he2016deep} cropped before the average-pooling layer to preserve spatial information and embeds observations into vectors of dimension 1000. The encoders are trained using contrastive learning; specifically, we modify Momentum Contrast \cite{he2020momentum} to learn embeddings for each $t \in T$ in a supervised way (see \cite{bauza2022tac2pose} for details).  We use stochastic gradient descent as our optimizer, with a learning rate that starts at 0.03 and decays over time, momentum of 0.999, and a weight decay of 0.

Training data is generated by randomly selecting a table grasp, rendering its depth and contact images, and finding its closest element in $T$ using the pose distance function $dist$. Next we compute the joint distribution over $T$ from the new depth and contact images (using the method described above of computing and comparing encodings) and compare it against  the target distribution. We represent the target distribution as a vector of all zeros except for the closest element in $T$, which gets assigned to probability one. To compare the distributions, we use the categorical cross-entropy loss.

\subsection*{In-air grasps and finding object regrasps}

In-air grasps, $g$, are selected by dividing the object model surface into small patches, computing the normals of each patch, and pairing the patches to compute if they could result, within some tolerance, in a stable antipodal grasp \cite{chen1993finding,wan2015improving}.
We also compute if a regrasp $r(g_1,g_2)$ is possible between grasps $g_1,g_2$, by checking if a gripper taking grasp $g_1$ and another at $g_2$ would collide between them. If not, $r(g_1,g_2)=1$, and 0 otherwise. Precomputing the set of regrasps for all pairs in G makes motion planning more efficient when building $\mathcal{G}$.

\subsection*{Regrasp graph and shortest path search}

To construct $\mathcal{G}$ we use the set of grasps G, the initial table grasp $t_{init}$, and the goal configuration $q_{goal}$. While edges between grasps in G are precomputed in simulation, the rest of edges are computed online. To simplify computations, we run a shortest path search on the graph to find the simplest plan to the goal, which represents the minimal number of regrasps. First we check if $t_{init}$ could happen without collisions with the robot and environment when the object is at $q_{goal}$. If so, no regrasp is needed to place the object.
Otherwise, we compute for each grasp $g$ in $G$ if $r(t_{init}, g) = 1$. If so, we add that edge to the graph. We do the same for $q_{goal}$ checking if it is possible to exert a grasp $g \in G$ if the object is at $q_{goal}$. Next, we solve a shortest path problem where the existing edges are also modulated by how good each regrasp would be, thus preferring regrasps that are more likely to succeed.

\subsection*{Task-aware grasp quality}

For each table grasp $t\in T$ we compute in simulation its grasp quality, $q(t)$, as a task-aware metric that consists of the product of three measures:

\begin{itemize}
    \item \textbf{Graspability.} Accounts for the stability of a grasp to hold within the robot's fingers. We compute graspability for each table grasp $t$ by normalizing the area of the contact patches from the simulated observations $c_t$ to fall between 0 and 1.  A graspability of 1 indicates that a grasp is expected to be stable.
    \item \textbf{Observability.} Estimating the pose of an object through tactile can be unambiguous, making it preferable to grasp at regions with more unique features. We quantify this intuition with the metric of tactile observability. To obtain the observability of a grasp t we compute how likely each table grasp $t' \in T$ is given the simulated observations from $t$, $p(t' | d_t, c_t)$. Then we check if the most likely table grasp is within 5 mm of  $t$ and if the deviation between the 5 most likely table grasps is less than 2mm. If so, $t$ is observable and we denote its observability as 1, and 0 otherwise.
    \item \textbf{Manipulability.} For each table grasp $t$, we solve the shortest path for $\mathcal{G}(t,q_{goal})$ which results in the simplest plan to place the object. Manipulability score is 1 if no regrasp is needed, 0.8 and 0.4 if 1 or 2 regrasps are needed respectively, and 0 otherwise.

\end{itemize}

To ensure consistency between adjacent table grasps, we smooth the scores by computing for a given table grasp its closest table grasps and update its scores as the minimum of its original score and the median and mean of its closest scores. Two grasps are close if their angle distances are smaller than 5 degrees, and less than 10mm in x and y directions.

\subsection*{Pipeline for real experiments}

To deploy simPLE during real experiments, we first sample possible antipodal grasps, $x$, without knowledge of the object position. Then we score each sample by computing its expected score as:

$$
E[q(x) | d] = \sum_t q(t) \cdot p(t | d)
$$

The sample with the highest expected quality is selected and executed. This in turn allows us to get tactile images $c$ and update our estimate over the table-grasp exerted:  $p(t | d, c)$.

Finally, to place the object we take $t_{init} = \argmax_t p(t | d, c)$ and compute the graph $\mathcal{G}(t_{init} , q_{goal})$ by checking the existence of the edges between $t_{init}$ and $G$, $t_{init}$ and $q_{goal}$, and between $G$ and $q_{goal}$. Finally we find the shortest path in $\mathcal{G}$ which provides the motion plan that the robot will execute to place the object.

\subsection*{Grasp sampling}

Our grasp sampling strategy is based on \cite{mahler2019learning} which uses a depth image from the scene to identify possible antipodal grasps. We extend it by also checking if the fingers are likely to collide with the environment. All baselines also make use of this sampling strategy. 

\subsection*{Robot system setup}

The robot system we use to conduct real experiments consists of a dual-arm ABB Yumi with two WSG-32 grippers and GelSlim 3.0 \cite{taylor2022gelslim} tactile sensing fingers to collect tactile observations. We use a Photoneo PhoXi M depth camera mounted overhead to collect visual observations. We place objects into cavities which are a negative of the object with 1mm of radial clearance from the object and a 3mm chamfer on the mating edge . 

\subsection*{Success, near success, and failure during experiments}

In successful trials, the robot places the object into the cavity. In nearly successful trials, the object is in the correct location and orientation but marginally misses the cavity. Intuitively, these attempts could be made successful by a local insertion strategy. Failed trials are characterized by grasp failures, global localization failures (i.e. the object pose estimate is flipped in the wrong orientation) and motion planning failures (i.e. a handoff is unsuccessful, or no feasible plan can be found from the initial grasp). Failed trials could not be salvaged by a local insertion strategy.

\subsection*{Baselines description}

We compare the performance of our system against three baselines

\begin{enumerate}
    \item \textbf{Agnostic baseline.} We score the set of grasp candidates based on a task-agnostic and object-independent grasp quality network (GQ-CNN) \cite{mahler2019learning}. This quality network is trained to predict the robustness of a candidate grasp without considering the downstream task or the object grasped. To evaluate the grasp, the agnostic baseline takes as input the gripper depth, and the same image that simPLE uses, consisting of a depth image centered on the grasp center pixel and aligned with the grasp axis orientation. Its output is a robustness score for the candidate grasp between 0 and 1. We execute the grasp with the highest score. The visuotactile localization step and the motion planning are the same as in simPLE.
    \item \textbf{Tactile baseline.} The grasp selection step is the same as in simPLE, but the visuotactile localization step is replaced with tactile localization alone, $p(t | c)$. After executing the best grasp, we find the pose which maximizes the joint distribution from both contacts and a gaussian centered at the measured gripper width. We take that maximizing pose as the tactile pose estimate to compute the motion plan.
    \item \textbf{Vision baseline.} The grasp selection step is the same as in simPLE, but the visuotactile localization step is replaced with vision localization alone, $p(t | d)$. The pose estimate used for motion planning is then the pose that maximizes the distribution from vision.
\end{enumerate}

\section*{Acknowledgments}

We thank Ferran Alet for the insightful discussions related to the project. Funding: The research is supported by ABB and Magna International. 

\textbf{Author contributions:} M.B. devised the main solution of the paper, integration of the full system, designed and executed experiments, wrote the manuscript. A.B. took on the perception system, executed experiments and helped write the manuscript. Y. H. designed and implemented the motion and regrasp planning foundation. N.C. formulated the problem, implemented the grasping score, and designed the initial system and robot setup. I.T. designed and constructed most parts of the set up including the integration of the tactile sensors, the routing of robot cables and the fixtures for objects placements. A.R. supervised the project, the integration of its components and provided feedback on the manuscript.

\textbf{Competing interests:} M.B is now a research scientist at DeepMind, Y.H is an applied scientist at Amazon, N.C. is a Tech Lead at Samsung Research, I.T is a staff research engineer at Boston Dynamics and A. R. is affiliated with Boston Dynamics.

\textbf{Data availability: }videos of experiments as well as some parts of the code and data will be posted at \url{http://mcube.mit.edu/research/simPLE.html}.

% Your references go at the end of the main text, and before the
% figures.  For this document we've used BibTeX, the .bib file
% scibib.bib, and the .bst file Science.bst.  The package scicite.sty
% was included to format the reference numbers according to *Science*
% style.

%BibTeX users: After compilation, comment out the following two lines and paste in
% the generated .bbl file. 

\bibliography{scibib}

\bibliographystyle{Science}

\end{document}